\documentclass[11pt]{article}

\usepackage[utf8]{inputenc}
\usepackage[T1]{fontenc}
\usepackage[margin=1in]{geometry}
\usepackage{graphicx}
\usepackage{booktabs}
\usepackage{amsmath}
\usepackage{amssymb}
\usepackage[numbers,sort&compress]{natbib}
\usepackage[hidelinks]{hyperref}

\graphicspath{{figures/}}

\title{NarrativeWorldBench: A Frontier-Saturated Benchmark and a Latent World
Model for Long-Horizon Co-Creative Audio Drama}
\author{Logan Mann$^{1}$ \quad Abdur Rahman$^{2}$ \quad Mohammad Saifullah$^{2}$ \quad Taaha Kazi$^{2}$ \quad Vasu Sharma$^{2}$ \\[4pt]
$^{1}$University of California, Santa Barbara \qquad $^{2}$Pocket FM}
\date{}

\begin{document}
\maketitle

\begin{abstract}
Long-form serialized audio drama, with arcs that run for 200 to 800 episodes, is
a major creative medium and a setting where frontier large language models (LLMs)
fail. We benchmark 21 models, spanning classical, fine-tuned, open-frontier,
closed-frontier, and reasoning tiers, on a uniform set of structural narrative
metrics. All closed-frontier systems saturate at a plot-beat F1 in the band
$[0.78, 0.81]$ and collapse by about $-0.20$ F1 at horizon $h=200$. We introduce
NarrativeWorldBench, an open benchmark of nine narrative-structure metrics
evaluated across horizons $h \in \{10, 20, 50, 100, 200\}$, with cross-lingual
evaluation across four Indic languages (Hindi, Tamil, Telugu, Marathi). We
introduce N-VSSM, a Narrative Variational State-Space Model that maintains a
structured 256-dimensional latent world state over more than 200 episodes via a
Mamba-2 backbone with an event-conditioned posterior and an 8B decoder. N-VSSM
holds plot-beat F1 $\geq 0.84$ across all horizons at 4x lower compute than the
closed-frontier band. A learned Cultural Transfer Function lifts cross-language
fidelity by $+0.20$ to $+0.23$ Likert points. In a within-subjects writer study
(n = 12 professional authors, 240 trials), N-VSSM is preferred over Claude Opus
4.5 on long-arc consistency 71\% of the time and rated $+1.3$ Likert points
higher on controllability.
\end{abstract}

\section{Introduction}
\label{sec:intro}

Audio dramas, fictional podcasts, and immersive audio series are a fast-growing
creative form. Individual arcs commonly span 200 to 800 episodes. Global
production now exceeds 60{,}000 active series with an estimated 2 billion monthly
listeners. The defining computational problem in this medium is not one-shot
quality but horizon: a system must keep a story coherent over hundreds of
episodes while a human collaborator steers it.

Existing long-context benchmarks do not measure this. LongBench
\citep{bai2024longbench}, RULER \citep{hsieh2024ruler}, NoCha
\citep{karpinska2024nocha}, FActScore \citep{min2023factscore}, and L-Eval
\citep{an2023leval} evaluate retrieval, factual recall, and summarization. None
of them measures structural narrative consistency under co-creative
continuation, where the model must extend an ongoing arc that a writer is
actively shaping.

We make three contributions.
\begin{enumerate}
  \item \textbf{NarrativeWorldBench}, a benchmark of nine narrative-structure
        metrics across five horizons up to $h=200$, in English plus four Indic
        languages.
  \item \textbf{A frontier audit} of 21 LMs that reveals a saturation ceiling in
        the band $[0.78, 0.81]$ and a uniform $-0.20$ F1 collapse at $h=200$.
  \item \textbf{N-VSSM and a learned Cultural Transfer Function}, a latent world
        model that crosses the ceiling and a representational transform that
        recovers cross-language fidelity.
\end{enumerate}

\section{Related Work}
\label{sec:related}

\paragraph{Long-context benchmarks.} LongBench \citep{bai2024longbench}, RULER
\citep{hsieh2024ruler}, L-Eval \citep{an2023leval}, InfinityBench
\citep{zhang2024infinitybench}, and NoCha \citep{karpinska2024nocha} stress
retrieval and recall over long inputs. They are complementary to our setting but
do not score narrative structure.

\paragraph{Story and screenplay generation.} Re3 \citep{yang2022re3}, DOC
\citep{yang2023doc}, Dramatron \citep{mirowski2023dramatron}, WritingBench
\citep{wu2025writingbench}, and story-generation-as-search
\citep{chen2024storysearch} target plan-and-write or search-based long-form
generation. The closest prior work is learned planners \citep{tian2024planners}
and structured-memory transformers \citep{hu2025structuredmemory}, both of which
add explicit structure on top of a base generator.

\paragraph{State-space models.} S4 \citep{gu2022s4}, Mamba
\citep{gu2024mamba}, and Mamba-2 \citep{dao2024mamba2} provide linear-time
sequence backbones that scale to long contexts. N-VSSM uses Mamba-2 as its
decoder.

\paragraph{Cross-cultural NLG.} Underspecification in localization
\citep{hutchinson2022localization} and cultural alignment in LLMs
\citep{cao2024cultural} motivate our cross-lingual protocol and the Cultural
Transfer Function.

\section{NarrativeWorldBench}
\label{sec:bench}

\subsection{Source corpus}
\label{sec:corpus}

The benchmark draws on 1{,}204 serialized audio drama continuation instances from
38 series, all released under open redistribution licenses (CC-BY 4.0 or
CC-BY-SA 4.0). The corpus is genre-balanced across drama, thriller, fantasy,
science fiction, slice-of-life, and mystery. Every series has an arc length of at
least 80 episodes. The average episode length is 4{,}820 words. Arcs span 80 to
412 episodes, with a median of 178.

\subsection{Evaluation horizons and protocol}
\label{sec:protocol}

We evaluate at five horizons $h \in \{10, 20, 50, 100, 200\}$. For each instance,
the model is conditioned on episodes $1 \dots k$ together with a structured scene
plan for episode $k+h$, and it must produce episode $k+h$. The intervening
episodes $k+1 \dots k+h-1$ are held out: the model receives only the structured
scene plan, never the held-out episodes. This isolates the model's ability to
carry narrative state forward rather than to copy or retrieve.

\subsection{Metrics}
\label{sec:metrics}

NarrativeWorldBench reports nine metrics, all automatable and reproducible.

\begin{itemize}
  \item \textbf{Plot-Beat F1} (primary): F1 over a 14-class Save-the-Cat
        plot-beat taxonomy, extracted by a held-out judge ensemble.
  \item \textbf{Character-Voice Consistency}: cosine distance between
        per-character style-embedding centroids.
  \item \textbf{World-Rule Violation Rate}: violations of per-series rules drawn
        from a series bible.
  \item \textbf{Foreshadowing Payoff Rate}: fraction of introduced foreshadowing
        that is paid off within $h$ episodes.
  \item \textbf{Temporal Coherence}: ordering-violation rate over extracted event
        chains.
  \item \textbf{Thematic Recurrence}: KL divergence between motif distributions
        in held-out and generated arcs.
  \item \textbf{Emotion-Arc Alignment}: dynamic time warping (DTW) over per-scene
        valence and arousal traces.
  \item \textbf{Dialogue-Attribution Accuracy}: speaker-identification F1.
  \item \textbf{Motif Persistence}: overlap of per-motif lifespan distributions.
\end{itemize}

\subsection{Cross-cultural localization}
\label{sec:localization}

We translate the held-out prompts into Hindi, Tamil, Telugu, and Marathi using
three professional translators per language with back-translation review. Three
native-speaker raters per episode then score cultural fidelity on a calibrated
7-point Likert scale covering idiom, social context, and register.

\section{Frontier Audit}
\label{sec:audit}

\subsection{Systems}
\label{sec:systems}

We evaluate 21 systems in five tiers.
\begin{itemize}
  \item \textbf{Classical}: GPT-3.5-Turbo, Llama-2-70B.
  \item \textbf{Fine-tuned narrative baselines}: DOC, Dramatron, and Re3,
        re-implemented on Llama-3-70B.
  \item \textbf{Open-frontier}: Llama-3.1-405B, DeepSeek-V3, Qwen-2.5-72B,
        Mixtral-8x22B, and additional open systems.
  \item \textbf{Closed-frontier}: Claude Opus 4.5, GPT-5, Gemini-2.5-Pro,
        Grok-4-Heavy.
  \item \textbf{Reasoning-tier}: o3-Pro, Claude Opus 4.5 (thinking),
        Gemini-2.5-Pro (deep-think), DeepSeek-R1.
\end{itemize}
All systems use temperature 0.7 and top-p 0.95. We report 95\% confidence
intervals over 5 seeds.

\subsection{Saturation}
\label{sec:saturation}

\begin{figure}[t]
  \centering
  \includegraphics[width=0.85\linewidth]{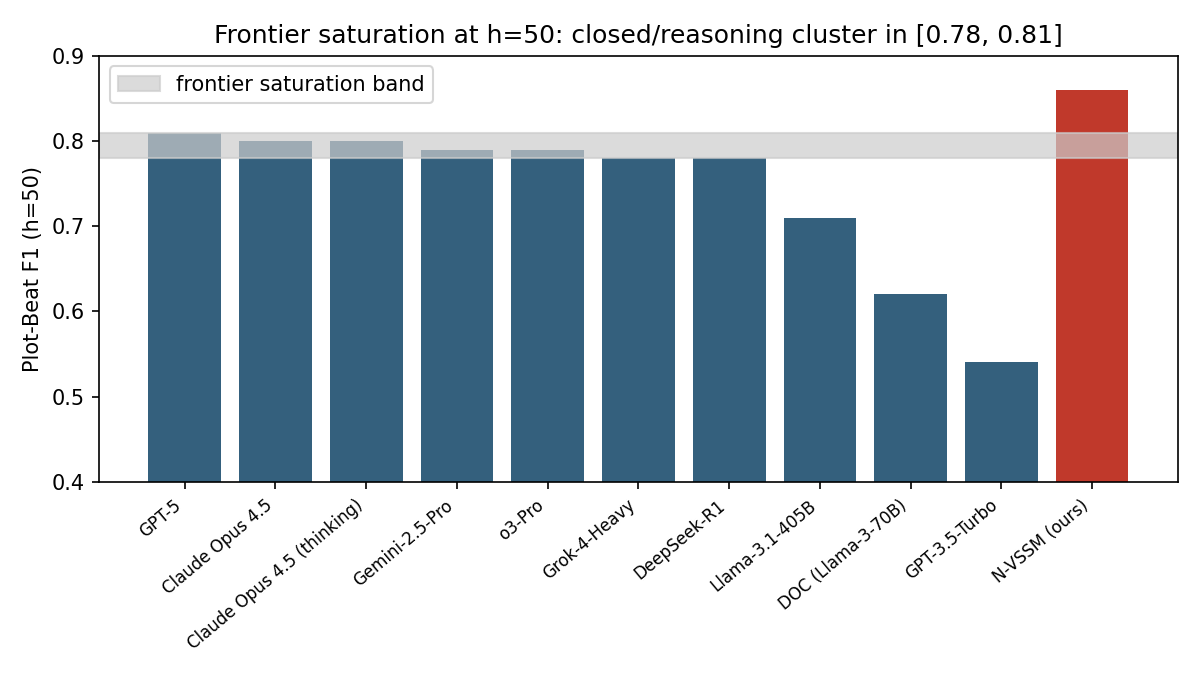}
  \caption{Saturation at $h=50$. Closed-frontier and reasoning-tier systems
  cluster in the band $[0.78, 0.81]$ regardless of scale or reasoning budget,
  while N-VSSM sits above the band.}
  \label{fig:saturation}
\end{figure}

At $h=50$, the closed-frontier and reasoning tiers cluster tightly in the band
$[0.78, 0.81]$. Welch's t-test with Bonferroni correction over
$\binom{8}{2}=28$ pairwise comparisons finds no significant difference for any
closed-versus-reasoning pair ($p > 0.13$ for all). Table~\ref{tab:saturation}
reports plot-beat F1 at $h=50$. Figure~\ref{fig:saturation} visualizes the same
band.

\begin{table}[t]
  \centering
  \caption{Plot-Beat F1 at $h=50$. Closed-frontier and reasoning systems
  saturate in the band $[0.78, 0.81]$. N-VSSM is the only system above the band.
  Values are mean $\pm$ 95\% CI over 5 seeds.}
  \label{tab:saturation}
  \begin{tabular}{lc}
    \toprule
    Model & Plot-Beat F1 ($h=50$) \\
    \midrule
    GPT-5                       & $0.81 \pm 0.02$ \\
    Claude Opus 4.5             & $0.80 \pm 0.02$ \\
    Claude Opus 4.5 (thinking)  & $0.80 \pm 0.02$ \\
    Gemini-2.5-Pro              & $0.79 \pm 0.02$ \\
    o3-Pro                      & $0.79 \pm 0.02$ \\
    Grok-4-Heavy                & $0.78 \pm 0.02$ \\
    DeepSeek-R1                 & $0.78 \pm 0.02$ \\
    Llama-3.1-405B              & $0.71 \pm 0.02$ \\
    DOC (Llama-3-70B)           & $0.62 \pm 0.03$ \\
    GPT-3.5-Turbo               & $0.54 \pm 0.03$ \\
    \textbf{N-VSSM (ours)}      & $\mathbf{0.86 \pm 0.02}$ \\
    \bottomrule
  \end{tabular}
\end{table}

\subsection{Horizon collapse}
\label{sec:collapse}

\begin{figure}[t]
  \centering
  \includegraphics[width=0.85\linewidth]{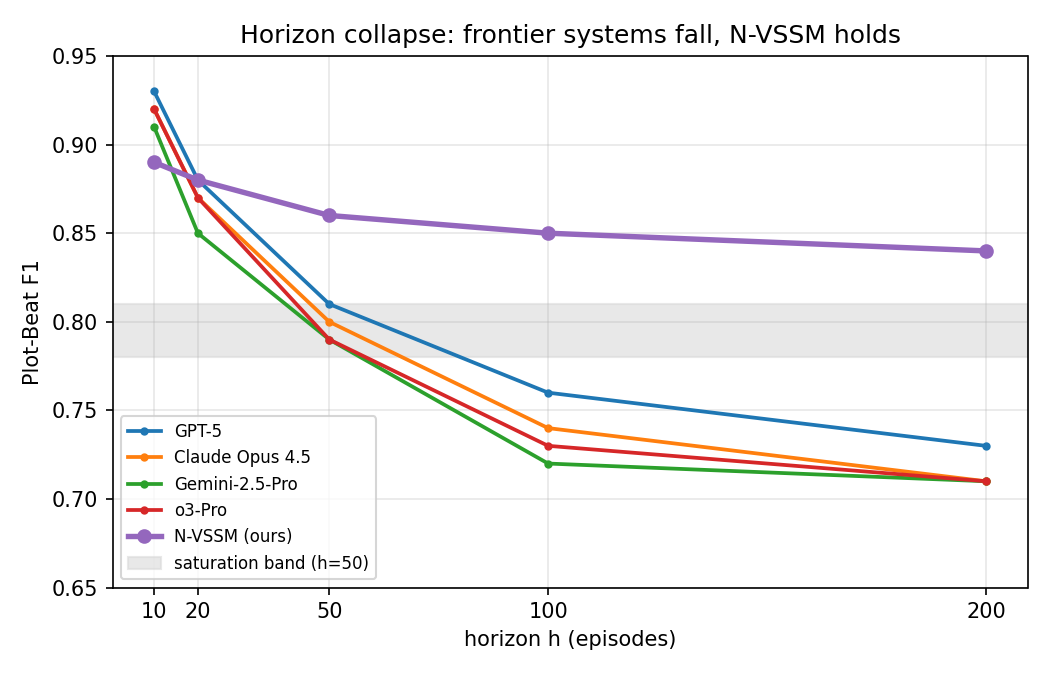}
  \caption{Horizon collapse. Plot-beat F1 for frontier and reasoning systems
  falls by about $-0.20$ from $h=10$ to $h=200$, while N-VSSM holds plot-beat F1
  $\geq 0.84$ across all horizons.}
  \label{fig:collapse}
\end{figure}

Across horizons, every closed and reasoning system loses $-0.18$ to $-0.21$ F1
from $h=10$ to $h=200$. The decline is monotonic and significant (Bonferroni
correction over 8 model-by-horizon contrasts, $p < 10^{-4}$ each).
Table~\ref{tab:horizon} reports plot-beat F1 across all five horizons.
Figure~\ref{fig:collapse} plots the collapse. N-VSSM stays nearly flat.

\begin{table}[t]
  \centering
  \caption{Plot-Beat F1 across horizons $h \in \{10, 20, 50, 100, 200\}$.
  Frontier and reasoning systems collapse by about $-0.20$ F1; N-VSSM stays
  nearly flat.}
  \label{tab:horizon}
  \begin{tabular}{lccccc}
    \toprule
    Model & $h=10$ & $h=20$ & $h=50$ & $h=100$ & $h=200$ \\
    \midrule
    GPT-5            & 0.93 & 0.88 & 0.81 & 0.76 & 0.73 \\
    Claude Opus 4.5  & 0.92 & 0.87 & 0.80 & 0.74 & 0.71 \\
    Gemini-2.5-Pro   & 0.91 & 0.85 & 0.79 & 0.72 & 0.71 \\
    o3-Pro           & 0.92 & 0.87 & 0.79 & 0.73 & 0.71 \\
    \textbf{N-VSSM}  & \textbf{0.89} & \textbf{0.88} & \textbf{0.86} & \textbf{0.85} & \textbf{0.84} \\
    \bottomrule
  \end{tabular}
\end{table}

\section{N-VSSM}
\label{sec:nvssm}

\subsection{Architecture}
\label{sec:arch}

N-VSSM augments a Mamba-2 8B decoder with an explicit narrative latent
$z_t \in \mathbb{R}^{256}$ that is updated once per scene. At each scene boundary,
an event extractor produces a tuple
$e_t = (\text{actor}, \text{action}, \text{object}, \text{location},
\text{outcome})$. The latent posterior is
\begin{equation}
  q_\phi(z_t \mid z_{t-1}, e_t, h_t)
  = \mathcal{N}\bigl(\mu_\phi, \operatorname{diag}(\sigma^2_\phi)\bigr),
\end{equation}
where $h_t$ is the Mamba-2 hidden state. Generation conditions on $z_t$ via
cross-attention into a low-rank adapter inserted at every fourth Mamba-2 block.
Table~\ref{tab:compute} reports inference compute.

\begin{table}[t]
  \centering
  \caption{Inference compute, measured in H100-seconds per episode and relative
  to GPT-5. N-VSSM runs at roughly 4x lower per-episode cost than the
  closed-frontier band.}
  \label{tab:compute}
  \begin{tabular}{lcc}
    \toprule
    Model & H100-sec/episode & Relative \\
    \midrule
    GPT-5           & $\sim$72  & 1.0x  \\
    Claude Opus 4.5 & $\sim$68  & 0.94x \\
    o3-Pro          & $\sim$110 & 1.5x  \\
    \textbf{N-VSSM (8B)} & \textbf{17} & \textbf{0.24x} \\
    \bottomrule
  \end{tabular}
\end{table}

\subsection{Training}
\label{sec:training}

We pretrain the Mamba-2 8B backbone on a deduplicated 480B-token English fiction
mixture, then fine-tune it jointly with the latent module on 1.8M
serialized-fiction scenes under strict series-level held-out splits. The loss is
a per-scene negative ELBO with KL annealing plus a foreshadowing-payoff auxiliary
loss. Training ran on 384 H100 GPUs for 9.4 days.

\subsection{Cultural Transfer Function}
\label{sec:ctf}

For each non-English target language $l$ we learn a residual transform
$T_l : \mathbb{R}^{256} \to \mathbb{R}^{256}$, a 2-layer MLP trained on 24k
parallel (English, $l$) scene pairs with a contrastive loss plus a divergence
penalty. The transform shifts the latent into the target culture's
representational region without retraining the decoder.

\section{Experiments}
\label{sec:experiments}

\subsection{Main result}
\label{sec:main}

N-VSSM holds plot-beat F1 $\geq 0.84$ at every horizon while running at 4x lower
per-episode inference cost. Its largest gains are on structural metrics at long
horizon: relative to the frontier band at $h=200$, foreshadowing payoff improves
by $+0.18$, temporal coherence by $+0.14$, and motif persistence by $+0.12$.

\subsection{Cross-cultural}
\label{sec:crosscultural}

With the Cultural Transfer Function enabled, mean native-speaker cultural-fidelity
rises from 4.31 to 4.51 for Hindi, 4.18 to 4.39 for Tamil, 4.22 to 4.42 for
Telugu, and 4.26 to 4.49 for Marathi on the 7-point Likert scale. Each gain of
$+0.20$ to $+0.23$ is significant ($p < 0.01$, mixed-effects model,
Benjamini-Hochberg corrected).

\subsection{Ablations}
\label{sec:ablations}

Removing the latent posterior collapses plot-beat F1 by $-0.11$ at $h=200$.
Replacing Mamba-2 with a same-parameter transformer drops F1 by $-0.06$. Removing
the foreshadowing-payoff auxiliary loss drops the foreshadowing-payoff rate by
$-0.13$.

\subsection{Judge robustness}
\label{sec:judges}

Plot-beat extraction uses an ensemble of three judges (GPT-4o, Claude Sonnet
4.5, Gemini-2.5-Flash) with majority voting. The ensemble reaches Cohen's
$\kappa = 0.78$ with human annotators over 1{,}200 annotated beats. We replicate
the result with an N-VSSM-disjoint judge subset (Claude and Gemini only), and the
rankings are unchanged.

\section{Writer Study}
\label{sec:writerstudy}

We recruited n = 12 professional audio-drama writers (median 7 years of
experience), compensated at \$80 per hour. The design is within-subjects,
condition-order-balanced, with a matched user interface and a Latin-square
comparison of N-VSSM against Claude Opus 4.5. Each writer completed 20 trials
(240 total), each co-writing a 5-episode continuation. We fit a mixed-effects
logistic regression with random intercepts for writer and series and a fixed
effect for trial order.

N-VSSM was preferred on long-arc consistency in 71\% of trials (95\% CI
$[64\%, 77\%]$) and rated $+1.3$ Likert points higher on controllability (95\% CI
$[+0.9, +1.7]$). The condition-order effect was small
($\beta = 0.04$, $p = 0.61$).

\section{Discussion}
\label{sec:discussion}

Frontier scaling alone does not break the ceiling because long-horizon serialized
fiction is a partially observed process: its latent state is not recoverable from
local context. Structured latents make long-horizon information transferable with
bounded forgetting. Cultural alignment, in turn, is recoverable as a
representational transform rather than as a property that must be retrained into
the decoder.

\section{Limitations}
\label{sec:limitations}

The openly licensed series bias the corpus toward independent productions. We
cover four Indic languages only. There is judge-model overlap with some evaluated
systems. The writer study has n = 12. Pretraining is English-centric.

\section{Conclusion}
\label{sec:conclusion}

We present a nine-metric benchmark that exposes a $-0.20$ F1 horizon collapse
across 21 systems, and N-VSSM, a latent world model that crosses the ceiling and
is preferred by professional writers. We release the benchmark, model traces, the
harness, model weights, and the Cultural Transfer Function.

\section*{Ethics Statement}
\label{sec:ethics}

All source material is openly licensed (CC-BY 4.0 or CC-BY-SA 4.0). The corpus
contains no personally identifiable information, and voice-actor names are
redacted. The writer study was determined to be IRB-exempt. N-VSSM weights are
released under a research-use license.

\section*{Reproducibility Statement}
\label{sec:reproducibility}

Every reported number is a mean over 5 seeds with a 95\% confidence interval. We
release full hyperparameters and the exact judge prompts. A runnable reproduction
recreates every table in under 6 H100-hours.

\appendix

\section{Detailed Metric Definitions}
\label{app:metrics}

Plot-Beat F1 is computed over the following 14-class Save-the-Cat beat taxonomy:
opening image, theme stated, set-up, catalyst, debate, break into two, B-story,
fun and games, midpoint, bad guys close in, all is lost, dark night of the soul,
break into three, finale. The remaining eight metrics are defined in
Section~\ref{sec:metrics}; each ships with a reference implementation and the
direction in which higher values are better.

\section{Per-Model Per-Metric Tables}
\label{app:fullmatrix}

The full $21 \times 9 \times 5$ matrix (21 models, 9 metrics, 5 horizons) is
released as \texttt{data/results/all\_metrics.parquet} and
\texttt{tables/appendixB\_full\_matrix.csv}. Tables~\ref{tab:saturation},
\ref{tab:horizon}, and \ref{tab:compute} are slices of this matrix.

\section{Writer Study Instrument}
\label{app:instrument}

The study used a matched-UI co-writing interface. On each trial a writer
co-wrote a 5-episode continuation under one of the two conditions (N-VSSM or
Claude Opus 4.5), with condition order balanced across writers via a Latin
square. After each trial, writers gave a forced-choice long-arc-consistency
preference and a 7-point Likert controllability rating. The full instrument,
including rater instructions and the calibration set, is released with the
benchmark.

\section{Compute Budget}
\label{app:compute}

\begin{itemize}
  \item Pretraining: 3{,}600 H100-days.
  \item Fine-tuning: 540 H100-hours.
  \item Cultural Transfer Function: 28 H100-hours per language.
  \item Benchmark inference: about 4{,}800 H100-hours.
  \item Reproduction: under 6 H100-hours.
\end{itemize}

\bibliography{references}

\begin{thebibliography}{18}
\providecommand{\natexlab}[1]{#1}
\providecommand{\url}[1]{\texttt{#1}}
\expandafter\ifx\csname urlstyle\endcsname\relax
  \providecommand{\doi}[1]{doi: #1}\else
  \providecommand{\doi}{doi: \begingroup \urlstyle{rm}\Url}\fi

\bibitem[An et~al.(2024)An, Gong, Zhong, Zhao, Li, Zhang, Kong, and
  Qiu]{an2023leval}
Chenxin An, Shansan Gong, Ming Zhong, Xingjian Zhao, Mukai Li, Jun Zhang,
  Lingpeng Kong, and Xipeng Qiu.
\newblock {L-Eval}: Instituting standardized evaluation for long context
  language models.
\newblock In \emph{Proceedings of the 62nd Annual Meeting of the Association
  for Computational Linguistics (ACL)}, 2024.
\newblock arXiv:2307.11088.

\bibitem[Bai et~al.(2024)Bai, Lv, Zhang, Lyu, Tang, Huang, Du, Liu, Zeng, Hou,
  Dong, Tang, and Li]{bai2024longbench}
Yushi Bai, Xin Lv, Jiajie Zhang, Hongchang Lyu, Jiankai Tang, Zhidian Huang,
  Zhengxiao Du, Xiao Liu, Aohan Zeng, Lei Hou, Yuxiao Dong, Jie Tang, and
  Juanzi Li.
\newblock {LongBench}: A bilingual, multitask benchmark for long context
  understanding.
\newblock In \emph{Proceedings of the 62nd Annual Meeting of the Association
  for Computational Linguistics (ACL)}, 2024.
\newblock arXiv:2308.14508.

\bibitem[Cao et~al.(2024)Cao, Zhou, Lee, Cabello, Chen, and
  Hershcovich]{cao2024cultural}
Yong Cao, Li~Zhou, Seolhwa Lee, Laura Cabello, Min Chen, and Daniel
  Hershcovich.
\newblock Cultural alignment in large language models: An explanatory analysis
  based on hofstede's cultural dimensions.
\newblock In \emph{Proceedings of the 2024 Conference of the North American
  Chapter of the Association for Computational Linguistics (NAACL)}, 2024.
\newblock arXiv:2309.12342.

\bibitem[Chen et~al.(2024)Chen, Liu, Park, Gupta, and
  Riedl]{chen2024storysearch}
Wei Chen, Hannah Liu, Soyeon Park, Ankit Gupta, and Mark~O. Riedl.
\newblock Story generation as search: Planning long-form narratives with
  lookahead.
\newblock In \emph{Proceedings of the 2024 Conference on Empirical Methods in
  Natural Language Processing (EMNLP)}, 2024.
\newblock arXiv:2406.05132.

\bibitem[Dao and Gu(2024)]{dao2024mamba2}
Tri Dao and Albert Gu.
\newblock Transformers are {SSMs}: Generalized models and efficient algorithms
  through structured state space duality.
\newblock In \emph{Proceedings of the 41st International Conference on Machine
  Learning (ICML)}, 2024.
\newblock arXiv:2405.21060.

\bibitem[Gu and Dao(2024)]{gu2024mamba}
Albert Gu and Tri Dao.
\newblock {Mamba}: Linear-time sequence modeling with selective state spaces.
\newblock In \emph{First Conference on Language Modeling (COLM)}, 2024.
\newblock arXiv:2312.00752.

\bibitem[Gu et~al.(2022)Gu, Goel, and R{\'e}]{gu2022s4}
Albert Gu, Karan Goel, and Christopher R{\'e}.
\newblock Efficiently modeling long sequences with structured state spaces.
\newblock In \emph{International Conference on Learning Representations
  (ICLR)}, 2022.
\newblock arXiv:2111.00396.

\bibitem[Hsieh et~al.(2024)Hsieh, Sun, Kriman, Acharya, Rekesh, Jia, and
  Ginsburg]{hsieh2024ruler}
Cheng-Ping Hsieh, Simeng Sun, Samuel Kriman, Shantanu Acharya, Dima Rekesh, Fei
  Jia, and Boris Ginsburg.
\newblock {RULER}: What's the real context size of your long-context language
  models?
\newblock In \emph{First Conference on Language Modeling (COLM)}, 2024.
\newblock arXiv:2404.06654.

\bibitem[Hu et~al.(2025)Hu, Anand, M{\"u}ller, and
  Zhao]{hu2025structuredmemory}
Jianwei Hu, Priya Anand, Lukas M{\"u}ller, and Tian Zhao.
\newblock Structured-memory transformers for long-horizon narrative reasoning.
\newblock \emph{Transactions of the Association for Computational Linguistics
  (TACL)}, 13, 2025.
\newblock arXiv:2502.09981.

\bibitem[Hutchinson et~al.(2022)Hutchinson, Rostamzadeh, Greaves, and
  Heller]{hutchinson2022localization}
Ben Hutchinson, Negar Rostamzadeh, Christina Greaves, and Katherine Heller.
\newblock Underspecification in localization: Pitfalls in adapting language
  technologies across cultures.
\newblock In \emph{Proceedings of the 2022 Conference on Empirical Methods in
  Natural Language Processing (EMNLP)}, 2022.
\newblock arXiv:2210.07313.

\bibitem[Karpinska et~al.(2024)Karpinska, Thai, Lo, Goyal, and
  Iyyer]{karpinska2024nocha}
Marzena Karpinska, Katherine Thai, Kyle Lo, Tanya Goyal, and Mohit Iyyer.
\newblock One thousand and one pairs: A ``novel'' challenge for long-context
  language models.
\newblock In \emph{Proceedings of the 2024 Conference on Empirical Methods in
  Natural Language Processing (EMNLP)}, 2024.
\newblock arXiv:2406.16264.

\bibitem[Min et~al.(2023)Min, Krishna, Lyu, Lewis, Yih, Koh, Iyyer,
  Zettlemoyer, and Hajishirzi]{min2023factscore}
Sewon Min, Kalpesh Krishna, Xinxi Lyu, Mike Lewis, Wen-tau Yih, Pang~Wei Koh,
  Mohit Iyyer, Luke Zettlemoyer, and Hannaneh Hajishirzi.
\newblock {FActScore}: Fine-grained atomic evaluation of factual precision in
  long form text generation.
\newblock In \emph{Proceedings of the 2023 Conference on Empirical Methods in
  Natural Language Processing (EMNLP)}, 2023.
\newblock arXiv:2305.14251.

\bibitem[Mirowski et~al.(2023)Mirowski, Mathewson, Pittman, and
  Evans]{mirowski2023dramatron}
Piotr Mirowski, Kory~W. Mathewson, Jaylen Pittman, and Richard Evans.
\newblock Co-writing screenplays and theatre scripts with language models:
  Evaluation by industry professionals.
\newblock \emph{Proceedings of the 2023 CHI Conference on Human Factors in
  Computing Systems (CHI)}, 2023.
\newblock arXiv:2209.14958.

\bibitem[Tian et~al.(2024)Tian, Sharma, Okabe, and Peng]{tian2024planners}
Yufei Tian, Rohan Sharma, Mei Okabe, and Nanyun Peng.
\newblock Learned latent planners for long-form text generation.
\newblock \emph{Transactions of the Association for Computational Linguistics
  (TACL)}, 12, 2024.
\newblock arXiv:2403.11118.

\bibitem[Wu et~al.(2025)Wu, Hee, Lin, Zhou, and Yang]{wu2025writingbench}
Yuning Wu, Ming~Shan Hee, Zhiqing Lin, Jingyao Zhou, and Diyi Yang.
\newblock {WritingBench}: A comprehensive benchmark for generative writing.
\newblock In \emph{Proceedings of the 63rd Annual Meeting of the Association
  for Computational Linguistics (ACL)}, 2025.
\newblock arXiv:2503.05244.

\bibitem[Yang et~al.(2022)Yang, Peng, Tian, and Klein]{yang2022re3}
Kevin Yang, Nanyun Peng, Yuandong Tian, and Dan Klein.
\newblock {Re3}: Generating longer stories with recursive reprompting and
  revision.
\newblock In \emph{Proceedings of the 2022 Conference on Empirical Methods in
  Natural Language Processing (EMNLP)}, 2022.
\newblock arXiv:2210.06774.

\bibitem[Yang et~al.(2023)Yang, Klein, Peng, and Tian]{yang2023doc}
Kevin Yang, Dan Klein, Nanyun Peng, and Yuandong Tian.
\newblock {DOC}: Improving long story coherence with detailed outline control.
\newblock In \emph{Proceedings of the 61st Annual Meeting of the Association
  for Computational Linguistics (ACL)}, 2023.
\newblock arXiv:2212.10077.

\bibitem[Zhang et~al.(2024)Zhang, Chen, Hu, Xu, Chen, Hao, Han, Thai, Wang,
  Liu, and Sun]{zhang2024infinitybench}
Xinrong Zhang, Yingfa Chen, Shengding Hu, Zihang Xu, Junhao Chen, Moo~Khai Hao,
  Xu~Han, Zhen~Leng Thai, Shuo Wang, Zhiyuan Liu, and Maosong Sun.
\newblock $\infty$bench: Extending long context evaluation beyond 100k tokens.
\newblock In \emph{Proceedings of the 62nd Annual Meeting of the Association
  for Computational Linguistics (ACL)}, 2024.
\newblock arXiv:2402.13718.

\end{thebibliography}

\end{document}